\documentclass{article}
\usepackage{spconf,amsmath,graphicx,hyperref}
\usepackage{amssymb}
\usepackage{booktabs}
\usepackage[table]{xcolor}

\title{RaFD: Flow-Guided Radar Detection for Robust Autonomous Driving}
\name{Shuocheng Yang, Zikun Xu, Jiahao Wang, Shahid Nawaz, Jianqiang Wang$^*$, Shaobing Xu$^*$
\thanks{This work was supported by the National Natural Science Foundation of China (No. 52372415, 52221005).}
\thanks{$^*$Corresponding author: \{shaobxu, wjqlws\}@tsinghua.edu.cn}
}
\address{School of Vehicle and Mobility\\
        Tsinghua University\\
        Beijing, China}
\begin{document}
%
\maketitle
\begin{abstract}
Radar has shown strong potential for robust perception in autonomous driving; however, raw radar images are frequently degraded by noise and “ghost” artifacts, making object detection based solely on semantic features highly challenging. To address this limitation, we introduce \textbf{RaFD}, a radar-based object detection framework that estimates inter-frame bird’s-eye-view (BEV) flow and leverages the resulting geometric cues to enhance detection accuracy. Specifically, we design a supervised flow estimation auxiliary task that is jointly trained with the detection network. The estimated flow is further utilized to guide feature propagation from the previous frame to the current one. Our flow-guided, radar-only detector achieves achieves \textbf{state-of-the-art performance} on the RADIATE dataset, underscoring the importance of incorporating geometric information to effectively interpret radar signals, which are inherently ambiguous in semantics.
\end{abstract}
\section{Introduction}
\label{sec:intro}

Current autonomous driving perception systems rely heavily on cameras and LiDAR~\cite{liBEVFormerLearningBirdsEyeView2022,philionLiftSplatShoot2020a,zhuCylindricalAsymmetrical3D2021}, yet their reliability degrades significantly under adverse weather (e.g., rain, snow, fog). Radar, with strong penetration capability, is far more robust in such conditions, making it a highly promising modality for achieving robust perception in autonomous driving.

However, raw radar signals are noisy, producing numerous “ghost” artifacts. In addition, compared with LiDAR, radar has lower azimuth resolution, leading to blurred object appearances. Consequently, single-frame radar perception~\cite{meyerGraphConvolutionalNetworks2021, popovNVRadarNetRealTimeRadar2023, bangRadarDistillBoostingRadarBased2024} suffers from limited semantic richness, posing significant challenges for object detection. 

\begin{figure}[t]
    \centering
    \includegraphics[width=\linewidth]{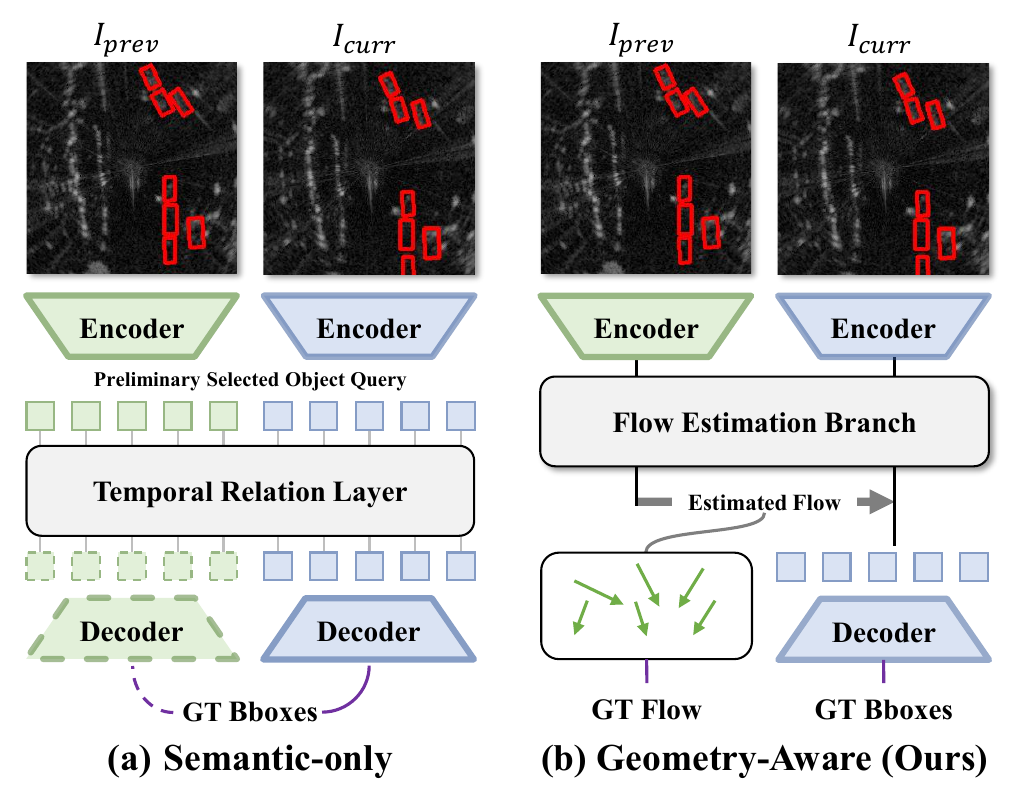}
    \caption{\textbf{Comparison between detection frameworks.} (a) The semantic-only framework~\cite{liExploitingTemporalRelations2022, yatakaRadarPerceptionScalable2024a, yatakaSIRAScalableInterFrame2024a} focuses solely on mining semantic features from consecutive frames. (b) Our geometry-aware framework explicitly incorporates geometric consistency through a flow estimation branch, where the estimated flow guides feature propagation to enhance detection.}
    \label{fig:intro}
\end{figure}

\begin{figure*}[t]
    \centering
    \includegraphics[width=\linewidth]{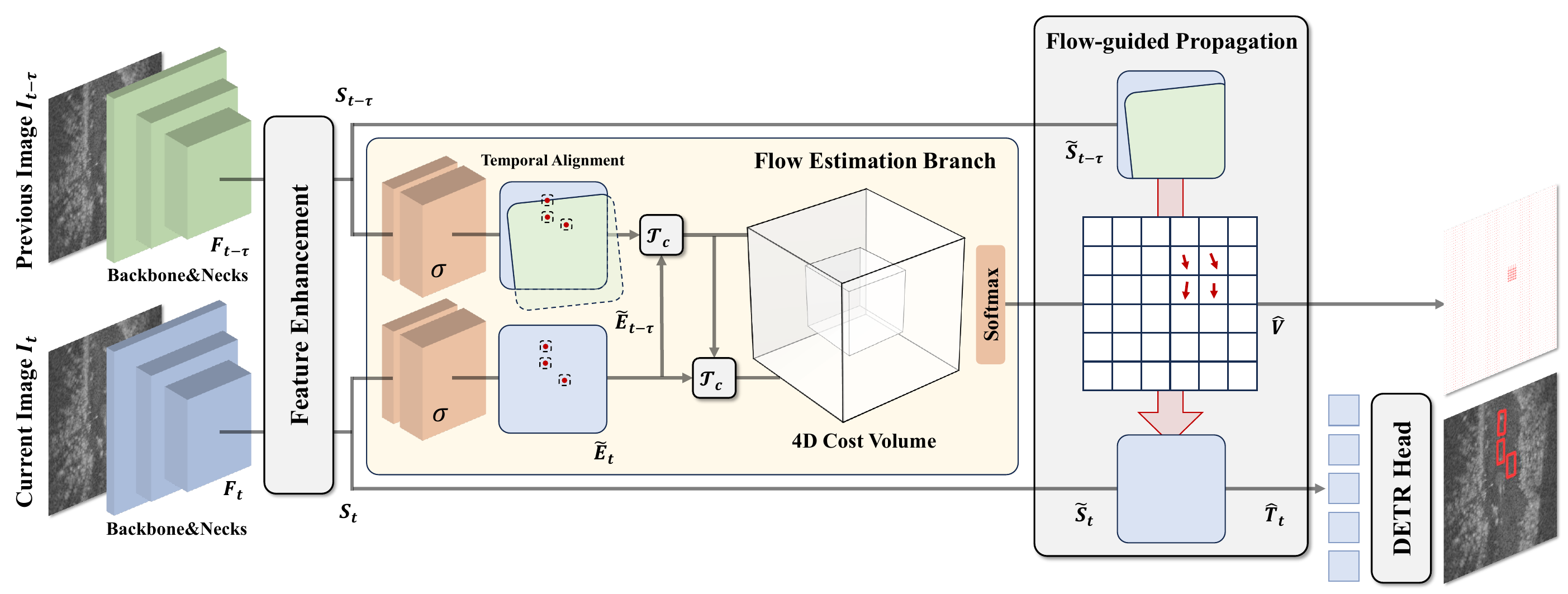}
    \caption{\textbf{The framework of the RaFD} (illustrated with two input frames).}
    \label{fig:overview}
\end{figure*}

To address this issue, several methods have attempted to leverage temporal information. TempoRadar~\cite{liExploitingTemporalRelations2022} introduced a temporal relation layer to associate potential object queries across two consecutive frames, while SIRA~\cite{yatakaSIRAScalableInterFrame2024a} extended this approach to multi-frame sequences. Despite promising results on RADIATE~\cite{sheenyRADIATERadarDataset2021} dataset, these methods rely solely on semantic cues. But in weak semantic signal interpretation, humans typically rely on motion patterns first—recognizing coherent trajectories—before reasoning about semantics. Thus we argue that \textbf{temporal radar detection should similarly prioritize cross-frame geometric consistency as a more reliable early-stage prior.}

Motivated by this, we propose RaFD, a \textbf{Ra}dar-based \textbf{F}low-guided \textbf{D}etector for robust autonomous driving. As illustrated in Fig.~\ref{fig:intro}, RaFD differs from previous radar sequence perception methods~\cite{liExploitingTemporalRelations2022, yatakaRadarPerceptionScalable2024a, yatakaSIRAScalableInterFrame2024a} by explicitly modeling the capture of geometric motion cues to enhance object detection. Specifically, inspired by advances in optical flow estimation~\cite{teedRAFTRecurrentAllPairs2020a, xuGMFlowLearningOptical2022}, we introduce a supervised auxiliary task—BEV scene flow estimation—that encourages the detector to learn temporally consistent geometric representations at the feature-map level. Notably, the ground-truth flow requires no additional labeling; we design an pipeline that automatically derives it from detection annotations with instance IDs. Building upon this, we further propose a feature propagation module that leverages the estimated flow to guide the aggregation of BEV features across frames for more robust representation learning. RaFD is structurally flexible, supporting seamless extension from two-frame input to multi-frame input. We validate RaFD on the RADIATE~\cite{sheenyRADIATERadarDataset2021} dataset under different input frame settings. \textbf{On the good-weather split, RaFD achieves 69.36\% mAP@0.3, 59.47\% mAP@0.5, and 23.92\% mAP@0.7, consistently outperforming existing state-of-the-art approaches.}
\section{Radar-based Flow-guided Detector}
\label{sec:rafd}

The overall framework of RaFD is illustrated in Fig.\ref{fig:overview}. We focus on scanning radar, represented as a BEV grayscale image $I \in \mathbb{R}^{1 \times H \times W}$. 

RaFD is built upon CenterFormer\cite{zhouCenterFormerCenterBasedTransformer2022}. Let $t$ denote the current frame and $t-\tau$ the previous frame. After backbone and neck encoding, we obtain feature maps $F_{t-\tau}$ and $F_t$, which are further enhanced to capture global scene context, yielding $S_{t-\tau}$ and $S_t$. These enhanced features are passed through a flow estimation branch to model inter-frame motion. The estimated flow $\hat{V}$ then guides feature propagation from the aligned $S_{t-\tau}$ to $S_t$, producing the final motion-aware representation $\hat{T}_t$. A convolutional head is subsequently applied to $\hat{T}_t$ to generate a heatmap of object centers. The top-$K$ peaks are extracted from this heatmap and used as initial queries for the DETR head, which refines them to predict offsets $(\hat{o}_x, \hat{o}_y)$, sizes $(\hat{h}, \hat{w})$, and orientations $\hat{\theta}$. In the following sections, we provide a detailed description of the key modules in the pipeline.

\subsection{Feature Enhancement}
\label{subsec:enhancement}

Since radar images are inherently noisy and blurred, relying solely on object appearance is unreliable. Optical flow estimation tasks face a similar challenge, as motion videos often contain blur, and GMFlow~\cite{xuGMFlowLearningOptical2022} highlights that enhancing features with spatial context is essential. Inspired by this, we adopt a transformer-based module, where self-attention aggregates relations across spatial locations. For efficiency, we employ shifted-window local attention, similar to Swin-Transformer~\cite{liuSwinTransformerHierarchical2021}, with window size $\tfrac{H_f}{2}\times \tfrac{W_f}{2}$, and use single-head attention to reduce computational complexity. Formally, one block is defined as
\begin{equation}
    \hat{F}^l=\mathcal{T}_{\text{w}}\big(F^{l-1}\big),\ F^l=\mathcal{T}_{\text{sw}}\big(\hat{F}^l\big).
\end{equation}
where $T_w$ and $T_{sw}$ denote regular and shifted-window self-attention, respectively. By stacking two such blocks, we obtain $S_{t-\tau}$ and $S_t$, which are enhanced with spatial context.

\subsection{Flow Estimation}
\label{subsec:flow}

Flow Estimation serves as an auxiliary task in RaFD, estimating motion vector fields between adjacent radar image frames at the feature-map level. Supervised by ground-truth flow, this task equips the front-end feature encoder with the ability to capture geometric consistency across the scene, while the predicted flow further supports feature propagation for object detection.

Given the spatial enhanced feature $S$, the network $\varphi$ encodes it into flow features $E = \varphi(S) \in \mathbb{R}^{{C_f}/{2} \times H_f \times W_f}$ using two weight-shared convolutional layers with batch normalization. We then align $E_{t-\tau}$ and $E_t$ by mapping each grid’s center coordinates $\mathcal{P}_t$ to the previous frame using the ego-pose transformation $\mathrm{T}_{t \rightarrow (t-\tau)}$. Bilinear interpolation retrieves features at the transformed points, while out-of-range locations default to $E_t$, producing aligned features $\widetilde{E}_{t-\tau}$ and $\widetilde{E}_t$ that represent the same real-world locations. This alignment also enables non-object regions to be set to zero during ground-truth flow generation.

\begin{figure}
    \centering
    \includegraphics[width=\linewidth]{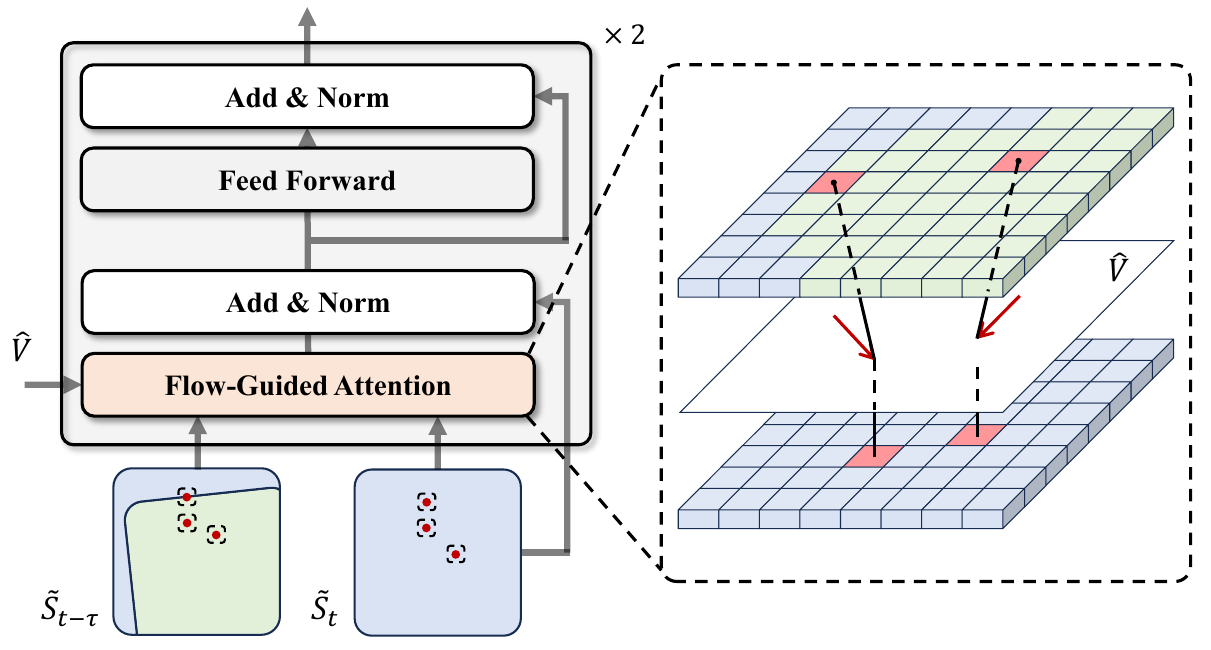}
    \caption{\textbf{Flow-Guided Attention.} The reference points in the deformable attention are adjusted according to the estimated flow. Red regions highlight object locations.} 
    \label{fig:fga}
\end{figure}

\begin{figure}
    \centering
    \includegraphics[width=\linewidth]{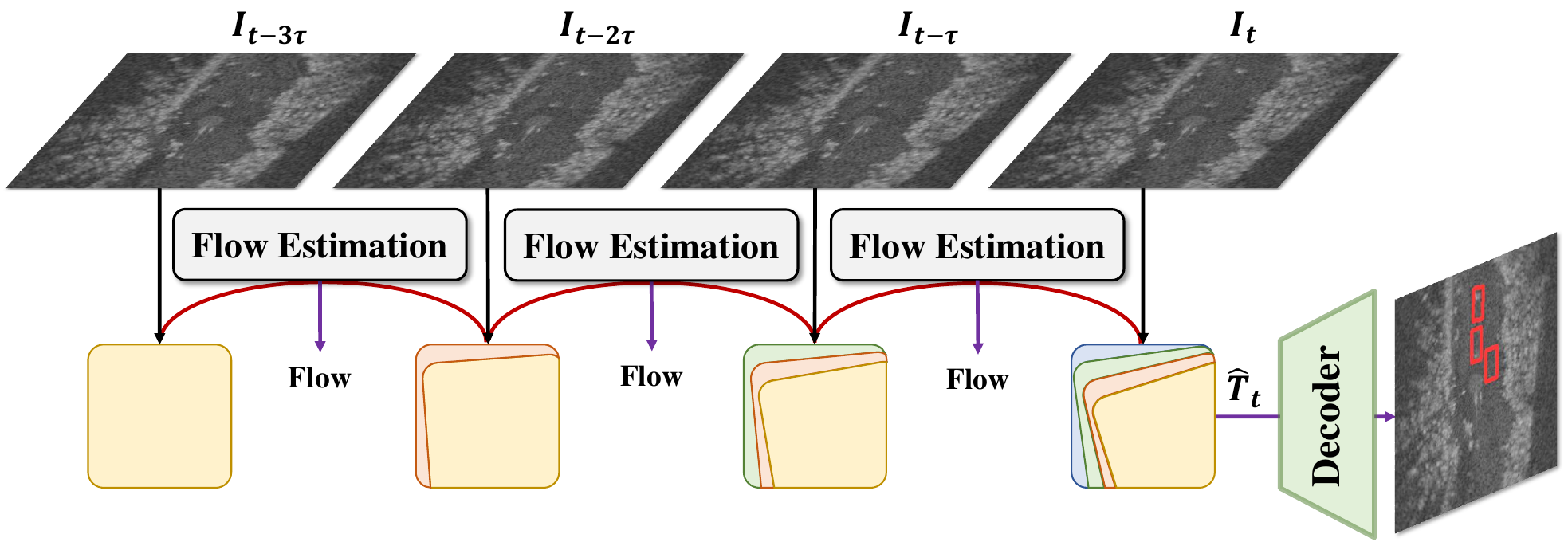}
    \caption{\textbf{The simple framework of RaFD with four-frame input.} Flow estimation is performed between every two consecutive frames. Several flow estimation and feature propagation modules share parameters.}
    \label{fig:ext}
\end{figure}

To capture the temporal dependency between $\widetilde{E}_{t-\tau}$ and $\widetilde{E}_t$, , we apply two global cross-attention blocks:
\begin{equation}
    \widetilde{E}_{t-\tau}^l=\mathcal{T}_{\text{c}}\big(\widetilde{E}_{t-\tau}^{l-1}, \widetilde{E}_t^{l-1}\big),\ 
    \widetilde{E}_t^l=\mathcal{T}_{\text{c}}\big(\widetilde{E}_t^{l-1},\hat{E}_{t-\tau}^l\big),
\end{equation}
where $\mathcal{T}_c$ denotes single-head cross-attention with feed-forward layers.

Next, we construct a 4D cost volume $C \in \mathbb{R}^{H \times W \times H \times W}$ via feature similarity,
\begin{equation}
    C^{(i,j,k,l)}=\frac{1}{\sqrt{C_f/2}}\sum_{c}^{C_f/2}\widetilde{E}_t^{(c,i,j)}\cdot\widetilde{E}_{t-\tau}^{(c,k,l)},
\end{equation}
where the brackets after the feature maps indicate the value at the given indices. This computation can be implemented efficiently with a simple matrix multiplication.

Selecting the highest similarity positions in $C$ would yield dense correspondences, but this operation is not differentiable. Inspired by GMFlow~\cite{xuGMFlowLearningOptical2022}, we instead normalize the last two dimensions of $C$ using a softmax operation to obtain a matching distribution. The flow $\hat{V}$ is then computed as:
\begin{equation}
    \hat{V}=G-G\cdot\text{softmax}(C)\in\mathbb{R}^{2\times H_f\times W_f}.
\end{equation}
where $G\in\mathbb{R}^{2\times H_f\times W_f}$ denote the 2D coordinates of the pixel grid on the feature map.

\subsection{Flow-guided Propagation}
\label{subsec:propagation}

The module is designed to propagate features from the previous frame to the current frame using the estimated flow $\hat{V}$. To achieve this, we adjust the reference points in deformable attention~\cite{zhuDeformableDETRDeformable2020} according to the flow, as shown in Fig.\ref{fig:fga}:
\begin{equation}
    \mathcal{R}_{t-\tau}=\big\{r^{(i,j)}_{t-\tau}=r^{(i,j)}_{t}-\hat{V}^{(i,j)}\big\}_{i=1,j=1}^{H_f,W_f},
\end{equation}
The flow-guided deformable attention then aggregates information from the temporally aligned features $\widetilde{S}_{t-\tau}$ and $\widetilde{S}_t$:
\begin{equation}
    \hat{T}_t=\mathcal{T}_{\text{f}}\big(\widetilde{S}_t,\widetilde{S}_{t-\tau}\big),
\end{equation}
where $\mathcal{T}_f$ denotes single-head flow-guided attention with feed-forward layers. We stack two such blocks to enable more comprehensive feature propagation.

\subsection{Extension}
\label{subsec:extension}

As shown in Fig.\ref{fig:ext}, RaFD can be naturally extended from two-frame to multi-frame inputs through its modular flow-guided propagation mechanism. For each consecutive pair of frames, the estimated flow vectors $\hat{V}_{t \to t-\tau}$ iteratively align and transfer features across the sequence, enabling long-range temporal propagation. This design maintains feature coherence over extended frame sequences and strengthens the model’s capacity to capture temporal dependencies in dynamic environments.

\subsection{Training}
\label{subsec:training}

Since the dataset does not provide scene flow annotations, we automatically derive pseudo ground-truth flow $V_{gt}$ from detection annotations with instance IDs. Specifically, $V_{gt}$ is constructed at the feature-map scale $H_f \times W_f$, where each object center is rendered as a Gaussian region with radius $\sigma=\max(f(hw), \gamma)$. With object IDs, instances are associated across frames, and the pose transformation $\mathrm{T}_{t \rightarrow (t-\tau)}$ is applied to align their coordinates. For each Gaussian region, the displacement of matched pixels defines the object flow, while background pixels are set to zero.

The predicted flow $\hat{V}$ is supervised using an $l_1$ loss:
\begin{equation}
\mathcal{L}_{\text{f}} = \lVert \hat{V} - V_{gt} \rVert_{1}.
\end{equation}

The overall training loss combines detection and flow supervision. 
\begin{equation}
\mathcal{L} = \mathcal{L}_{\text{det}}+\mathcal{L}_{\text{flow}}.
\end{equation}
We supervise the flow between every two consecutive input.
\begin{table*}
    \centering
    \setlength\tabcolsep{6pt}
    {
    \begin{tabular}{lccccccccc}
        \toprule
        
        & \multicolumn{4}{c}{Trained on good-weather split} & & \multicolumn{4}{c}{Trained on good-and-bad-weather split} \\
        \cmidrule{2-5}\cmidrule{7-10}
        & mAP@0.3 & mAP@0.5 & mAP@0.7 & EPE & & mAP@0.3 & mAP@0.5 & mAP@0.7 & EPE \\
        
        \midrule
        
        CenterPoint (1) & 59.42\scriptsize{$\pm$ 1.92} & 50.17\scriptsize{$\pm$ 1.91} & 18.93\scriptsize{$\pm$ 1.46} & - & & 53.92\scriptsize{$\pm$ 3.44} & 42.81\scriptsize{$\pm$ 3.04} & 13.43\scriptsize{$\pm$ 1.92} & - \\
        CenterFormer (1) & 61.79\scriptsize{$\pm$ 1.37} & 52.57\scriptsize{$\pm$ 1.53} & 19.24\scriptsize{$\pm$ 0.96} & - & & 57.13\scriptsize{$\pm$ 1.75} & 44.80\scriptsize{$\pm$ 1.32} & 14.55\scriptsize{$\pm$ 0.82} & - \\
        
        \midrule
    
        TempoRadar (2) & 63.63\scriptsize{$\pm$ 2.08} & 54.00\scriptsize{$\pm$ 2.16} & 21.08\scriptsize{$\pm$ 1.66} & - & & 56.18\scriptsize{$\pm$ 4.27} & 43.98\scriptsize{$\pm$ 3.75} & 14.35\scriptsize{$\pm$ 2.15} & - \\
        \textbf{RaFD (2)} & \cellcolor{gray!20}\textbf{65.89}\scriptsize{$\pm$ 0.97} & \cellcolor{gray!20}\textbf{55.13}\scriptsize{$\pm$ 1.26} & \cellcolor{gray!20}\textbf{22.75}\scriptsize{$\pm$ 1.03} & 0.1689 & & \cellcolor{gray!20}\textbf{61.95}\scriptsize{$\pm$ 2.07} & \cellcolor{gray!20}\textbf{50.23}\scriptsize{$\pm$ 1.83} & \cellcolor{gray!20}\textbf{17.58}\scriptsize{$\pm$ 1.44} & 0.1778 \\
    
        \midrule
    
        SCTR (4) & 68.06\scriptsize{$\pm$ 1.60} & 57.03\scriptsize{$\pm$ 1.34} & 22.62\scriptsize{$\pm$ 1.18} & - & & 66.01\scriptsize{$\pm$ 1.05} & 52.55\scriptsize{$\pm$ 0.96} & 19.18\scriptsize{$\pm$ 1.02} & - \\
        SIRA (4) & 68.68\scriptsize{$\pm$ 1.12} & 58.11\scriptsize{$\pm$ 1.40} & 22.81\scriptsize{$\pm$ 0.86} & - & & 66.14\scriptsize{$\pm$ 0.83} & 53.79\scriptsize{$\pm$ 1.14} & \textbf{19.85}\scriptsize{$\pm$ 0.95} & - \\
        \textbf{RaFD (4)} & \cellcolor{gray!20}\textbf{69.36}\scriptsize{$\pm$ 1.45} & \cellcolor{gray!20}\textbf{59.47}\scriptsize{$\pm$ 1.92} & \cellcolor{gray!20}\textbf{23.92}\scriptsize{$\pm$ 1.11} & 0.1556 & & \cellcolor{gray!20}\textbf{68.83}\scriptsize{$\pm$ 1.81} & \cellcolor{gray!20}\textbf{54.20}\scriptsize{$\pm$ 1.66} & 18.84\scriptsize{$\pm$ 1.56} & 0.1669 \\
    
        \bottomrule
    \end{tabular}
    }
    \caption{Benchmark results of different methods on the RADIATE dataset. Numbers in parentheses indicate frames input.}
    \label{tab:benchmark}
\end{table*}

\begin{table}[t]
    \footnotesize
    \centering
    \setlength\tabcolsep{4pt}
    {
    \begin{tabular}{lcccc}
        \toprule
         & mAP@0.3 & mAP@0.5 & mAP@0.7\\
        \midrule
        Baseline & 64.29 & 55.14 & 21.79 \\
        + Feature Enhancement & 66.92 (+2.63) & 56.80 (+1.66) & 22.41 (+0.64) \\
        + Flow Estimation & 67.76 (+0.84) & 57.78 (+0.98) & 22.69 (+0.28) \\
        + Flow-guided Propagation & 69.36 (+1.60) & 59.47 (+1.69) & 23.92 (+1.23) \\
        \bottomrule
    \end{tabular}
    }
    \caption{Ablation study of RaFD on RADIATE dataset. Baseline refers to four-frame input CenterFormer~\cite{zhouCenterFormerCenterBasedTransformer2022} using vanilla deformable attention~\cite{zhuDeformableDETRDeformable2020} for cross-frame feature propagation.}
    \label{tab:ablation}
\end{table}

\section{Experiment}
\label{sec:experiment}

\subsection{Experimental Setup}
\label{subsec:setup}

We evaluate RaFD on the RADIATE dataset~\cite{sheenyRADIATERadarDataset2021}, which provides high-resolution radar images (but without Doppler dimension) under diverse weather conditions. Following \cite{liExploitingTemporalRelations2022, yatakaRadarPerceptionScalable2024a, yatakaSIRAScalableInterFrame2024a}, we adopt the predefined train-on-good-weather, train-on-good-and-bad-weather, and test splits. Pedestrians are excluded from detection, and input images are center-cropped to $256\times256$, and objects outside this region are ignored. Relative poses between consecutive frames are obtained via~\cite{yangRINOAccurateRobust2025a}. Training is performed on 8 RTX 3090 GPUs with batch size 8, learning rate $2 \times 10^{-4}$, and weight decay $1 \times 10^{-2}$ using the Adam for 10 epochs. Oriented object detection is evaluated with mean Average Precision (mAP) at IoU thresholds 0.3, 0.5, 0.7, and flow estimation with End-Point Error (EPE), defined as the average $\ell_2$ distance between predicted and ground-truth flow vectors.

\subsection{Results Analysis}
\label{subsec:analysis}

\begin{figure}[t]
    \centering
    \includegraphics[width=\linewidth]{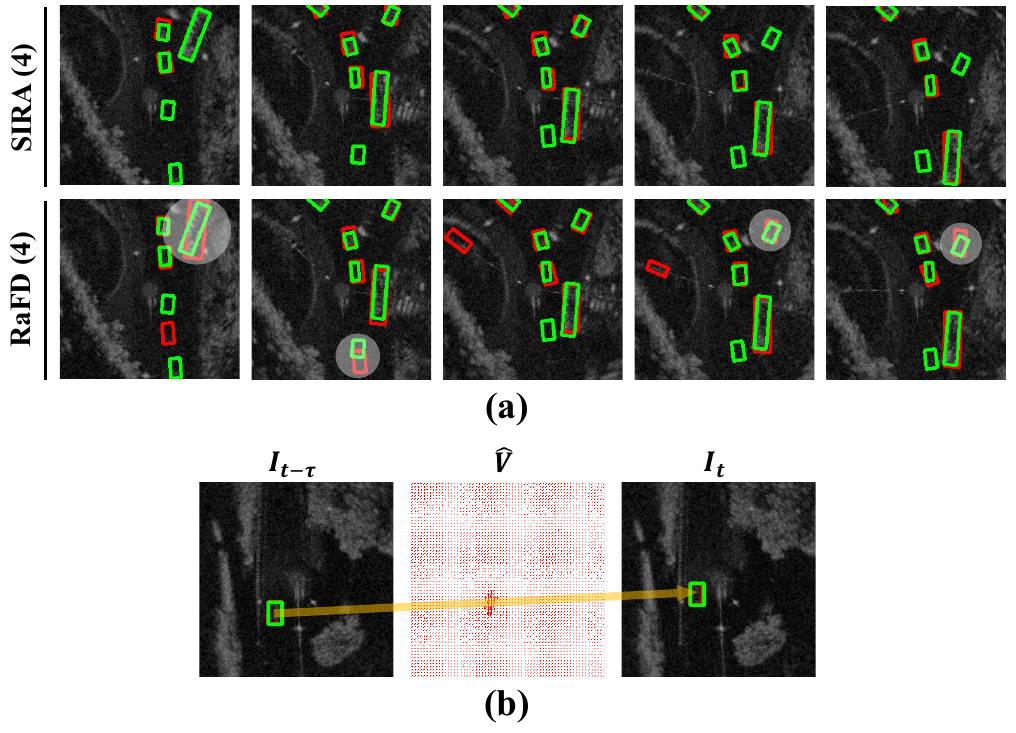}
    \caption{\textbf{Visualization of object detection and flow estimation}. Green boxes represent ground truth, while red boxes represent predictions.}
    \label{fig:visual}
\end{figure}

The quantitative results are summarized in Tab.\ref{tab:benchmark} and Tab.\ref{tab:ablation}. RaFD is compared with CenterPoint, CenterFormer, TempoRadar~\cite{liExploitingTemporalRelations2022}, SCTR~\cite{yatakaRadarPerceptionScalable2024a}, and SIRA~\cite{yatakaSIRAScalableInterFrame2024a}, all using ResNet-34 backbones. RaFD consistently achieves the best performance for both two-frame and four-frame inputs. Notably, the performance degradation on the good-and-bad-weather split is minimal compared to \cite{liExploitingTemporalRelations2022, yatakaRadarPerceptionScalable2024a, yatakaSIRAScalableInterFrame2024a}, which we attribute to the relatively stable performance of the flow estimation. This further demonstrates the beneficial effect of flow guidance for object detection. Incorporating feature enhancement, flow estimation, and flow-guided propagation modules leads to consistent performance gains, validating the effectiveness of each component. The visualization results are provided in Fig.\ref{fig:visual} for intuitive understanding.

\section{Conclusion}
\label{sec:conclusion}

In this paper, we present RaFD, a radar-based flow-guided detector designed for robust autonomous driving, which explicitly estimates and leverages inter-frame geometric clues. By incorporating an auxiliary supervised flow estimation task, RaFD captures temporal consistency and guides feature propagation across frames. Extensive experiments on the RADIATE dataset demonstrate that RaFD outperforms existing radar-only methods, offering a robust, accurate, and scalable solution for radar perception in challenging scenarios.

\newpage
\bibliographystyle{IEEEbib}
\bibliography{strings, main}

\end{document}